# QRetinex-Net: Quaternion-Valued Retinex Decomposition for Low-Level Computer Vision Applications

Sos Agaian, *Life Fellow, IEEE,* Vladimir Frants

*Abstract*—Images captured in low-light conditions often display considerable degradation, including color distortion, low contrast, noise, and artifacts, all of which reduce the accuracy of recognition tasks in computer vision. Retinex-based decomposition is commonly utilized in low-level vision tasks. Land and McCann's Retinex theory assumes that observed images can be decomposed into reflectance (R) and illumination (I) components: S = R ⊙ I, where ⊙ denotes element-wise multiplication. This decomposition leverages human ability to perceive object colors as relatively stable under varying illumination conditions. However, this decomposition is highly ill-posed, and designing proper constraint functions adaptive to various scenes remains challenging. Traditional Retinex decomposition models suffer from several fundamental limitations: (i) the original formulation assumes independence in processing the color channels (red, green, and blue), corresponding to the L, M, and S cones in the human visual system; (ii) while they decompose images into reflectance and illumination components, they lack neuroscientific accuracy in modeling human color perception; (iii) they do not guarantee perfect reconstruction of the original image; and (iv) they fail to adequately model how the human visual system achieves color constancy under varying illumination conditions. This study proposes a pioneering Quaternion Retinex decomposition, expressed as S = R ⊗ I, where ⊗ represents the Hamilton product. We introduce the Reflectance Consistency Index (RCI), a novel metric quantifying reflectance stability across illumination conditions. Extensive experiments on low-light crack detection, face detection under varying illumination, and infrared-visible fusion demonstrate our method's effectiveness. Results show superior color preservation, noise suppression, and reflectance consistency compared to conventional decomposition approaches, with 2-11% performance improvements across tasks.

*Index Terms* - Quaternions, Deep Learning, Retinex theory, Image Processing.

## I. INTRODUCTION

Image decomposition into the reflectance and illumination components has been proven beneficial in various computer vision tasks, ranging from object detection and tracking to face recognition in low-light conditions, where robust color constancy and accurate separation of intrinsic factors are crucial [1], [2], [3], [4], [5]. While Retinex theory remains foundational for illumination-reflectance decomposition, classical formulations diverge from modern neuroscientific understanding and often fail to ensure accurate signal reconstruction [6], [7]. These methods process color channels independently, ignoring the vectorial nature of color vision and cross-channel correlations [8]. Early approaches like Single-Scale Retinex (SSR) and Multi-Scale Retinex with Color Restoration (MSRCR) improved upon basic models through Gaussian or logarithmic filtering, but introduced halo artifacts, over-enhancement, and color shifts under challenging illumination [9][10]. Variational methods provide better stability through smoothness priors, yet may oversmooth details or amplify noise [11], [12]. Fusion-based strategies aggregate multiple enhancement estimates but struggle with large illumination variations [13]. Neuroscientific evidence reveals key limitations in classical Retinex frameworks. Spatial processing begins in early visual pathways (retina, LGN), where spatial frequency tuning and center-surround mechanisms drive coarse-to-fine processing [14]. However, Retinex lacks spatial-frequency modeling, decomposing images pixel-wise without considering broader spatial structure. Higher brain regions integrate color and spatial information jointly, with ventral areas like V4 showing enhanced responses to color boundaries [15].

This motivates quaternion-based representations that preserve inter-channel relationships through Hamilton product and color space rotations [16], [17], [18]. Hypercomplex algebra unifies color operations, reduces parameter redundancy, and shows promise for restoration, segmentation, and correction tasks [19]. Parallel research highlights the efficacy of wavelet transforms in handling both low- and high-frequency components, benefiting applications such as multi-scale feature extraction, denoising, and contrast enhancement [20], [21], [22]. Despite promising results, existing quaternion-wavelet methods often lack precise reconstruction guarantees or impose high computational demands, making them less suitable for tasks requiring exact pixel-level processing. So, despite their widespread adoption, Retinex decomposition models exhibit fundamental limitations, motivating our quaternion-based approach. Retinex decomposition is inherently ill-posed, leading to ambiguity and difficulty in obtaining a unique solution [23]. Traditional methods often produce halo artifacts and color distortions around high-contrast edges, negatively affecting perceptual quality [24], [25]. These models typically assume smooth illumination, inadequately representing scenes with complex lighting conditions, including shadows and multiple light sources [26], [27].

Vladimir Frants is with the Graduate Center, City University of New York, New York, NY 10016 USA (e-mail: vfrants@gradcenter.cuny.edu).
Sos Agaian is with the College of Staten Island, City University of New York, NY 10016 USA (e-mail: sos.agaian@csi.cuny.edu).

skip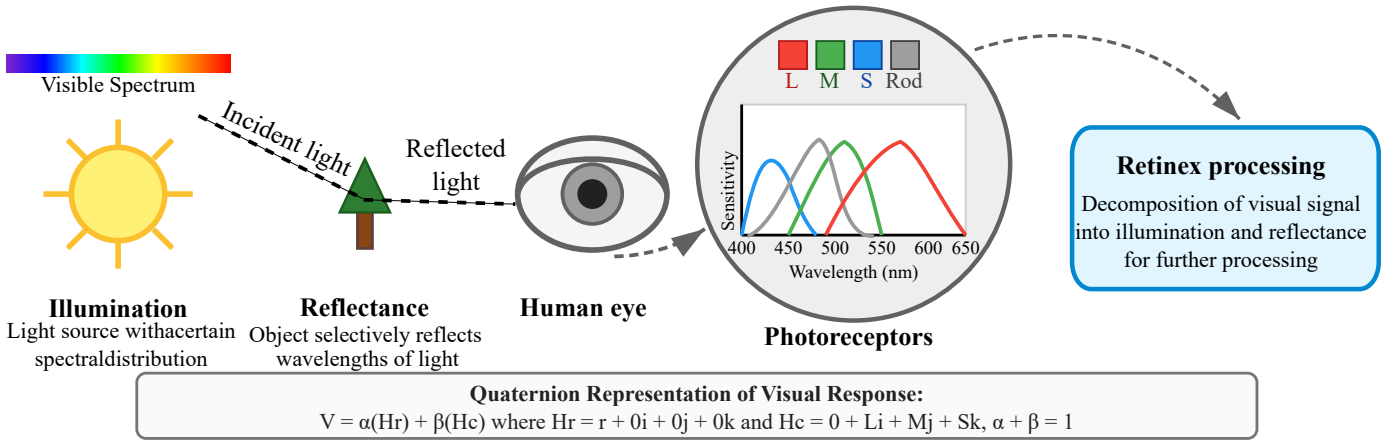

**Fig. 1.** Overview of the color vision pipeline and Retinex decomposition. An illumination source with a specific spectral distribution (left) illuminates an object whose surface reflectance selectively reflects certain wavelengths. The reflected light enters the human eye, where rods and three types of cones (L, M, and S) detect overlapping wavelength bands. Retinex processing then decomposes the visual signal into illumination and reflectance components for improved color constancy and image enhancement.

Additionally, Retinex techniques rely heavily on heuristic parameter tuning, resulting in inconsistent performance [28], and more advanced models frequently suffer from high computational complexity, limiting real-time usability [29].

They also amplify image noise, reducing effectiveness for noisy inputs [12], and perform poorly in extremely dark or saturated regions, creating unnatural artifacts [30]. Critically, traditional approaches process RGB channels independently, neglecting intrinsic color correlations crucial for accurate reproduction and color constancy [31]. Tackling these issues one by one often creates new problems, emphasizing the need for a fundamental improvement, like the quaternion-based framework introduced here [32].

This work proposes a novel wavelet-domain quaternion-based Retinex decomposition that ensures accurate reconstruction while robustly preserving color information. The main contributions are:

    1. Quaternion Retinex Framework (QRetinex-Net): We introduce a pioneering decomposition expressed as $S = R \otimes I$, where $\otimes$ the Hamilton product represents the operation in the wavelet domain to jointly model all color channels.

    2. Reflectance Consistency Index (RCI), which quantifies reflectance map stability under varying conditions.

    3. Multi-application validation across low-light crack detection, infrared-visible fusion, and face detection, demonstrating improved segmentation accuracy, detection performance, and fusion quality.

    4. Superior Performance on benchmarking against RetinexNet [33], KIND++ [34], U-RetinexNet [35], and Diff-Retinex [36] shows our method achieves better PSNR/SSIM scores, superior perceptual quality (LPIPS as low as 0.0001), and high reflectance consistency (RCI up to 0.988).

Paper Structure: Section II covers background on Retinex theory, quaternion algebra, and related work. Section III details our wavelet domain framework, network architecture, and RCI metric. Section IV presents experimental results, comparisons, and ablation studies. Section V concludes with future directions.

## II. BACKGROUND

### A. Retinex: Theory and Implementations

Vision is the ability to perceive light patterns from the external environment and gather information about the world from the images formed. The world we visually experience is filled with light and color; however, it is essential to recognize that color is not a physical property of the light itself. Although the wavelength or combination of wavelengths can characterize light, "color" is fundamentally a perceptual construct arising from complex processing within the human visual system and brain. Despite being a subjective experience, color perception often correlates strongly with the physical wavelengths of light, although this correlation is neither perfect nor direct.

Color perception depends on the interactions between illumination and the reflectance properties of the surfaces. As illustrated in Fig. 1, the process begins with an illumination source emitting light with a specific spectral distribution, which then interacts with object surfaces that selectively reflect specific wavelengths. This reflected light enters the human eye, where photoreceptors with distinct spectral sensitivities detect it. Different illuminant-surface combinations can produce identical color signals, leading to inherent ambiguity in the visual system. The overlapping spectral sensitivities of the cone cells in the human retina further intensify this ambiguity.

Humans typically possess three types of cones, categorized as L (long-wavelength), M (medium-wavelength), and S (short-wavelength) cones. The peak sensitivity of L-cones typically ranges between 564-580 nm, M-cones between 534-545 nm, and S-cones between 420-440 nm. While this overlapping sensitivity brings advantages for everyday vision, it significantly complicates accurate color reproduction. For instance, monochromatic yellow light can appear identical to a mixture of red and green light because both stimuli can trigger the same L and M cone responses, despite their distinct spectral compositions. This phenomenon is particularly problematic for narrow-band light sources, subtle color distinctions, or scenarios requiring high precision in color reproduction. Additionally, individual variations in cone responses mean that two observers may not perceive the same color under identical viewing conditions, further complicating the development of universal color reproduction systems. Retinex is considered one of the most influential approaches to understanding color constancy in particular and human vision in general, and has been applied to image enhancement [6], [7]. Retinex processing involves the decomposition of visual signals into illumination and reflectance components. The term "Retinex" itself is a combination of "retina" and "cortex," suggesting a processing pathway that spans from the initial detection of the light in the retina to higher-level analysis in the visual cortex. The theory proposes that our perception of color and lightness largely depends on factors beyond the specific illumination conditions, indicating its relevance in scenarios where illumination varies, yet objects' perceived colors and lightness remain relatively constant. In its original formulation, Retinex theory postulated a degree of independence in processing the color channels: red, green, and blue, corresponding to the L, M, and S cones in our eyes. This initial simplification assumed that each channel is processed in isolation, with each cone system calculating its lightness values without direct influence from the other channels.

Despite the elegant simplicity of the independent channel processing model, substantial evidence indicates that this conceptualization has significant limitations. Physiologically, our L and M cones show considerable overlap in spectral sensitivities, rendering true independence at the receptor level impossible. Moreover, extensive research supports cross-channel interactions in neural processing pathways beyond the retina, emphasizing the interconnectedness of color processing. Neuroscientific research has demonstrated that color constancy is processed by higher zones of the brain, particularly in area V4 of the visual cortex [37]. These higher processing centers maintain a hierarchical representation of color information, progressively abstracting color properties from the initial receptor inputs to more complex color constancy mechanisms. This hierarchical processing allows the visual system to maintain consistent color perception despite variations in illumination conditions, supporting the sophisticated color constancy capabilities observed in human vision. Perceptually, phenomena like simultaneous color contrast illustrate that surrounding colors affect our color perception, highlighting cross-channel influences. These perceptual effects emphasize the limitations of the independent channel model. The ambiguity arising from overlapping cone sensitivities is especially problematic for narrow-band light sources, subtle color differences, precise color reproduction in displays, and accounting for individual variations in cone cell responses.

The core idea of Retinex Theory is that the reflectance of a scene, the actual color of objects, and illumination are disentangled and processed separately. Mathematically, this is often expressed as:
$$S(x, y, \lambda) = R(x, y, \lambda) \odot I(x, y, \lambda)$$
where $S(x, y, \lambda)$ represents the observed image intensity at position $(x, y)$ and wavelength $\lambda$, $R(x, y, \lambda)$ represents the reflectance of the surface, and $I(x, y, \lambda)$ represents the illumination. In the context of digital images, Retinex-based methods attempt to enhance the image by compensating for variations in illumination, thereby improving perceived contrast and color fidelity.

Computer simulations of Retinex-based algorithms have demonstrated sensitivity to high-frequency components, enabling effective enhancement of image edge information [10]. Despite these advantages, early Retinex-based enhancements often appeared unnatural or over-enhanced, frequently resulting in color shifts, artifacts, and inadequate noise management, particularly in low-light conditions. Additionally, the edges in the enhanced images were sometimes not sharp enough, and high-frequency details could not be significantly improved. A detailed summary of these challenges is provided in [38].

The fundamental challenges in color constancy and Retinex-based approaches highlight the limitations of independent or straightforward channel processing. Complex visual tasks require more sophisticated color vision models that handle inherent ambiguities, illumination, and surface reflectance variations. Moreover, the ill-posed nature of the Retinex equation means infinitely many possible reflectance-illumination pairs can result in the same observed image. Key questions arise from this ill-posed condition: (i) What properties of surface collections are most relevant to real-world or industrial applications? (ii) Given that humans only have three types of cones, what can the visual system realistically infer about the actual reflectance of surfaces? And (iii) how can surface reflectance be reliably estimated from the light that enters the eye or camera sensor? Answering these questions requires improvements in color constancy algorithms, better human color perception modeling, and robust methods to manage cone sensitivity overlaps.

Novel mathematical approaches can help tackle the ill-posed Retinex equation and achieve color constancy challenges. Quaternion algebra presents a promising avenue, as it can represent both color and geometry within a unified framework. Quaternions provide a mathematical structure for color representation, allowing RGB color values to be modeled as a 3D vector component while using the scalar component for luminance information. This 4D representation naturally accommodates rotations and transformations in color space and can effectively model the spectral distribution and geometric relationships of color signals in a unified framework. The quaternion-based modeling of rod and cone cell relationships can be formulated through a comprehensive mathematical framework. A quaternion q can be represented as:
$$q = w + x\mathbf{i} + y\mathbf{j} + z\mathbf{k}$$
where w represents the rod response (scalar part), and $(x, y, z)$ represents the cone responses (vector part).

The quaternion representation of visual response can be expressed as:
$$V = \alpha(H_r) + \beta(H_c)$$
where the rod component is:
$$H_r = r + 0\mathbf{i} + 0\mathbf{j} + 0\mathbf{k}$$
with r being the scalar response magnitude of rods under given illumination conditions, and the cone component is:
$$H_c = 0 + L\mathbf{i} + M\mathbf{j} + S\mathbf{k}$$
where L, M, and S represent the responses of the three cone types. The parameters $\alpha$ and $\beta$ are light-dependent factors representing rod and cone contributions, respectively, with a normalization constraint of $\alpha + \beta = 1$.

This quaternion-based approach offers several key advantages: (i) Unified representation of rod and cone responses, natural handling of transitions between vision types, and efficient computation of spatial relationships, (ii) More accurate modeling of visual responses, better predictions of visual performance, and enhanced color image processing systems, and (iii) Examination of visual adaptation mechanisms, exploring rod-cone interactions, and advancing vision correction technologies. The framework can be extended to include additional factors such as chromatic adaptation, temporal dynamics, and spatial interactions, making it a versatile tool for both research and practical applications in vision science.

Early implementations of Retinex theory often processed the illumination channel through logarithmic or Gaussian filtering techniques. SSR used a center/surround filtering mechanism to enhance local contrast [9]. Although SSR offered improved color consistency, it could not simultaneously handle significant dynamic range compression and faithful tone rendering, frequently producing halo artifacts near edges. The MSRCR attempted to mitigate color distortions by integrating multiple scales and applying color recovery [10]. While MSRCR improved overall color fidelity relative to SSR, it still tended to introduce artifacts around high-contrast boundaries. Path-based Retinex algorithms explored pixel intensities along specific paths, but they proved computationally expensive, vulnerable to sampling noise, and highly dependent on path selection [28]. Subsequent classical variations, such as random spray Retinex (RSR) and bright-pass filtering, sought more efficient approximations of illumination. Nonetheless, these methods continued to struggle in the presence of strongly non-uniform illumination or intense noise, often resulting in over-enhancement or the loss of fine details [11], [12].

To address the inherent shortcomings of classical Retinex, variational-based frameworks introduced smoothness priors on illumination and piecewise continuity constraints on the reflectance map. Notable examples include total variation models (TVM) and weighted variational approaches (WV-SIRE), which demonstrated enhanced decomposition performance by enforcing regularity [39], [29]. However, these approaches sometimes over-smooth key edges and details in darker regions. Fusion-based methods similarly aimed to obtain better local detail enhancement by decomposing an image into reflectance and illumination and then combining multiple contrast-enhanced versions of the illumination map [13]. Despite improved detail restoration, these approaches could amplify noise or produce artifacts in scenes with drastic illumination differences. Lately, fractional-order Retinex approaches have leveraged fractional calculus to improve the conditioning of low-light decomposition and preserve complex textural details in under-exposed traffic images [40]. Meanwhile, plug-and-play Retinex strategies incorporating shrinkage mapping for illumination estimation have demonstrated enhanced noise suppression and brightness adjustment, further advancing the utility of Retinex-based frameworks [41]. With the advent of deep learning, Retinex-based decomposition progressed beyond hand-crafted constraints. RetinexNet introduced a two-stage pipeline that first separated the input into illumination and reflectance, then refined the illumination via a trained convolutional network [33]. However, it amplifies noise within the reflectance component in extremely low-light conditions.

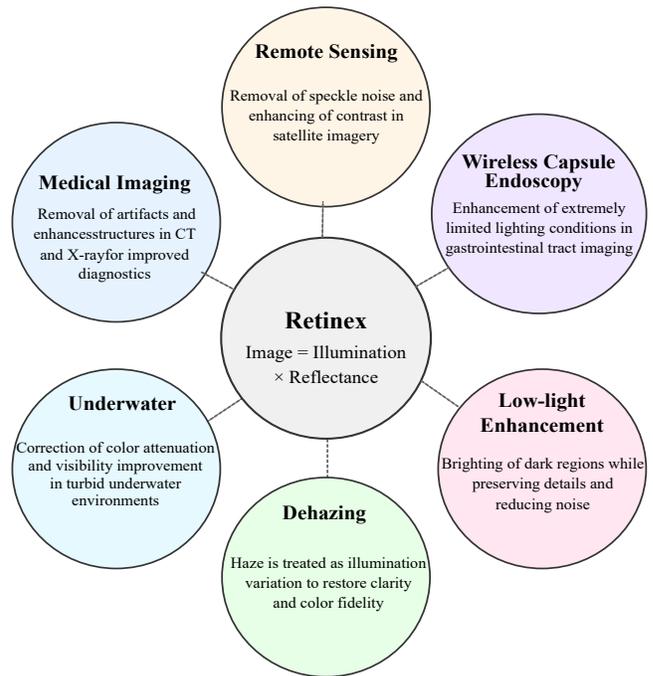

**Fig. 2.** Representative low-level vision applications leveraging the proposed quaternion-valued Retinex decomposition framework.

Building on this concept, KinD divided the workflow into decomposition, reflectance restoration, and illumination adjustment modules, aided by multi-scale illumination attention [42]. Although KinD improved global and local consistency, it tends to produce overexposed images under severe underexposure and requires careful tuning to balance denoising and detail preservation. Later architecture became more intricate. URetinex-Net mapped an iterative optimization process into a trainable network, refining both reflectance and illumination using learned priors [35]. Although this approach offered better noise handling, it demanded higher computation and reduced interpretability. Self-supervised or zero-shot frameworks, such as RetinexDIP [43] and RRDNet [44] allowed decomposition and enhancement without labeled training data, but often introduced additional complexity and longer processing times. Recent developments have explored transformer-based solutions. RetinexFormer integrates attention mechanisms to manage noise and uneven lighting while maintaining the traditional reflectance-illumination separation pipeline [45]. Further, hybrid methods for low-light image enhancement have combined diffusion or state-space models with Retinex ideas to improve generative capabilities and interpretability, at the cost of higher computational overhead [46], [36]. Retinex decomposition methods have evolved significantly from early single-scale filtering-based approaches to sophisticated deep-learning frameworks. Nevertheless, each category faces its limitations, ranging from halo artifacts and noise amplification to high computational costs and over-smoothing. The quaternion representation offers a promising direction for developing such solutions by providing a unified mathematical framework that can accommodate both the physiological reality of the human visual system and the computational requirements of modern image processing algorithms.

## B. Applications of Retinex

The ability of Retinex theory to separate intrinsic object properties makes it applicable across a range of imaging applications. By isolating reflectance and illumination, Retinex-based methods can help to mitigate various degradation factors, such as reflections, haze, and non-uniform illumination, while retaining crucial details. This subsection reviews key applications of Retinex decomposition in medical imaging, low-light endoscopy, dehazing, remote sensing, and underwater image enhancement. The summary of applications is given in Fig. 2.

*CT Metal Artifact Reduction:* In Computed Tomography (CT), metallic implants or instruments often create artifacts that obscure anatomical structures. Retinex-based techniques reduce these distortions by separating the metal-induced illumination component from the inherent tissue reflectance. For example, RetinexFlow integrates Retinex theory with a flow-based completion mechanism to remove metal artifacts while preserving critical features in reconstructed CT scans [47]. Similarly, Fu et al. proposed a refined Retinex-Net for X-ray enhancement, combining weighted guided filtering with local detail enhancement, to suppress noise and enhance contrast, thereby improving visual clarity [48].

**X-Ray Imaging:** Liu et al. introduced a single-exposure X-ray imaging method that filters out low-energy X-rays for high absorption ratio objects, effectively compressing the dynamic range and preventing saturation. This workflow subsequently applies a Retinex-based enhancement, using a global–local attention U-Net for illumination and an anisotropic diffused residual dense network for reflection, to restore critical structure information in final X-ray images [49].

**Wireless Capsule Endoscopy (WCE):** Although Retinex-based methods significantly improve quality of images in many low-light scenarios, WCE poses distinct challenges due to its extremely limited onboard lighting and the gastrointestinal (GI) tract's complex environment. Reduced illumination complicates the capture of sufficient visual details, often degrading the performance of classical methods. Conventional approaches such as gamma correction and histogram equalization offer partial improvements but falter under severe underexposure or intricate noise patterns. Mou et al. proposed a nonlinear luminance enhancement and denoising network that automatically refines brightness through higher-order curve functions and suppresses noise with a dual-attention mechanism, improving both visual clarity and quantitative metrics (PSNR, SSIM, LPIPS) [50]. Similarly, An et al. introduced EIEN, relying on a retrained Retinex-Net as a decomposition network alongside a self-attention guided multi-scale pyramid to correct illumination, then stretches the reflection component's color channels to amplify vascular and tissue information [51].

**Single Image Dehazing:** Retinex theory also is valuable for haze removal by treating haze primarily as an illumination variation. Gao et al. combined multi-scale Retinex with guided filtering and gamma correction to remove haze, retaining details and minimizing color distortion [52]. Li et al. introduced a deep Retinex Network (RDN) to learn residual illumination under hazy conditions, leveraging attention mechanisms for improved interpretability and generalization [53]. Other hybrid methods, such as ODD-Net, fuse classical Retinex concepts with advanced convolutional architectures to estimate atmospheric light and transmission more accurately, restoring both visibility and color fidelity [54].

**Remote Sensing:** Retinex decomposition helps reduce speckle noise and correct low contrast in satellite and aerial imagery. Shastry et al. employed a Retinex-based deep image prior (DIP) with a self-attention module to remove speckle noise from a single degraded satellite image without ground-truth data [55]. Liu et al. adapted Retinex-Net to multi-spectral remote sensing data, enhancing the intensity channel in IHS color space and improving multi-band contrast, however, further fine-tuning is needed for broader applicability [56].

**Underwater Image Enhancement:** Turbid underwater environments introduce color attenuation and low visibility. Zhang et al. proposed LAB-MSR, which applies bilateral and trilateral filtering in the CIELAB color space under a multi-scale Retinex framework, leading to improved contrast and color fidelity [57]. Li et al. introduced an adaptive weighted multi-scale Retinex (AWMR) approach that partitions an image into sub-blocks for refined scale selection [58]. The results are then fused in the gradient domain to prevent blocking artifacts.

Although Retinex-based decomposition methods have found diverse applications, each application poses unique challenges. Most recent methods do not reconstruct the separately processed components, due to the risk of introducing artifacts and suffering from limited reconstruction accuracy. In the applications mentioned above, Retinex is often used as the foundational approach. Addressing these issues is precisely the goal of our novel Quaternion Retinex decomposition.

## III. METHODOLOGY

In this section, we present our quaternion-valued Retinex decomposition framework. We first give a brief overview of quaternion color representation and introduce quaternion decomposition. Then, we detail our decomposition network architecture and its implementation details. Finally, we introduce the RCI metric.

### A. Quaternion Retinex

Quaternions extend complex numbers to four dimensions. A quaternion $q \in \mathbb{H}$ is expressed as $q = w + x\mathbf{i} + y\mathbf{j} + z\mathbf{k}$ where $w, x, y, z \in \mathbb{R}$ and the basis units satisfy: $\mathbf{i}^2 = \mathbf{j}^2 = \mathbf{k}^2 = -1$, $\mathbf{ij} = \mathbf{k}$, $\mathbf{ji} = -\mathbf{k}$, $\mathbf{jk} = \mathbf{i}$, $\mathbf{kj} = -\mathbf{i}$, $\mathbf{ki} = \mathbf{j}, \mathbf{ik} = -\mathbf{j}$. Multiplication on quaternions is defined through the Hamilton product. For two quaternions $q_1, q_2 \in \mathbb{H}$ the Hamilton product is defined as [16]:

$$q_1 \otimes q_2 = (w_1 w_2 - x_1 x_2 - y_1 y_2 - z_1 z_2) \\ + (w_1 x_2 + x_1 w_2 + y_1 z_2 - z_1 y_2)\mathbf{i} \\ + (w_1 y_2 - x_1 z_2 + y_1 w_2 + z_1 x_2)\mathbf{j} \\ + (w_1 z_2 + x_1 y_2 - y_1 x_2 + z_1 w_2)\mathbf{k} \quad (1)$$

Let $S \in \mathbb{H}^{H \times W}$ to be a quaternion-valued matrix representing an RGB image of height H and width W with color channels $R, G, B \in \mathbb{R}$ for red, green, and blue components, respectively.

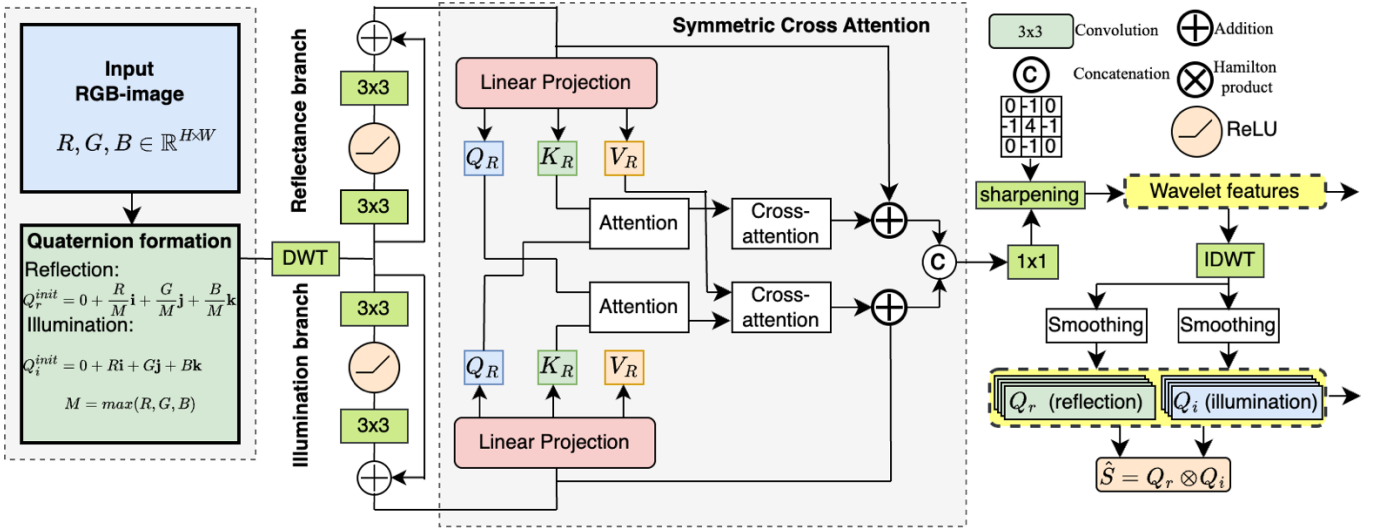

**Fig 3.** Quaternion-valued Retinex decomposition framework designed to (i) preserve inter-channel color relationships, (ii) ensure perfect reconstruction, and (iii) learn stable reflectance features across varying illuminations. The core idea is to represent color images as pairs of quaternions in the wavelet domain and to design a network that learns them so that their Hamilton product accurately reconstructs the original image.

Each element of S is a purely imaginary quaternion representing a single pixel:

$$S(x,y) = 0 + R(x,y)\boldsymbol{i} + G(x,y)\boldsymbol{j} + B(x,y)\boldsymbol{k} \quad (2)$$

where $1 \leq x \leq H, 1 \leq y \leq W$.

We aim to decompose S into two matrices $Q_R \in \mathbb{H}^{H \times W}$ and $Q_I \in \mathbb{H}^{H \times W}$ such that the element-wise Hamilton product yields reconstruction $\hat{S}$ close to S:

$$\hat{S}(x,y) = Q_R(x,y) \otimes Q_I(x,y) \quad (3)$$

Here, $Q_R$ is a reflectance-like component, invariant to the illumination changes, and $Q_I$ captures illumination information.

To facilitate the quaternion-based Retinex decomposition, we first express each RGB pixel as two quaternions $Q_R^{init}$ and $Q_I^{init}$. The reflectance-like map captures normalized color ratios that are invariant to the intensity:

$$Q_R^{init}(x,y) = 0 + \frac{R(x,y)}{M(x,y)}\boldsymbol{i} + \frac{G(x,y)}{M(x,y)}\boldsymbol{j} + \frac{B(x,y)}{M(x,y)}\boldsymbol{k} \quad (4)$$

where

$$M(x,y) = \max\{R(x,y), G(x,y), B(x,y)\} \quad (5)$$

The illumination-like map retains the original magnitudes:

$$Q_I^{init}(x,y) = 0 + R(x,y)\boldsymbol{i} + G(x,y)\boldsymbol{j} + B(x,y)\boldsymbol{k} \quad (6)$$

To exploit multi-scale information, we operate in the Haar wavelet domain [59]. Let $Q_R^{init} \in \mathbb{H}^{H \times W}$ and $Q_I^{init} \in \mathbb{H}^{H \times W}$ denote these initial reflectance and illumination quaternion maps. We apply the discrete Haar wavelet transform DWT to each channel of $Q_R^{init}$ and $Q_I^{init}$, yielding $\{Q_R^{init}, Q_I^{init}\} \mapsto \{Q_R^W, Q_I^W\}$, where $Q_R^W$ and $Q_I^W$ contain the low- and high-frequency sub-bands {LL, LH, HL, HH} per channel. A trainable decomposition network D then refines these wavelet-domain representations to produce $Q_R^{W*}$, $Q_I^{W*}$, which better disentangle the underlying reflectance structure from illumination. Finally, we invert the wavelet coefficients of each channel to obtain the outputs $Q_R$ and $Q_I$. The goal of the network is to obtain two quaternions whose Hamilton product reconstructs the input image and can be interpreted, respectively, as disentangled reflectance and illumination factors. This design preserves full reconstruction fidelity through the invertibility of the wavelet transform and the explicit quaternion splitting, while also leveraging inter-channel dependencies to improve color constancy and robust feature extraction.

*B. Decomposition Network*

The proposed quaternion decomposition network D generates the wavelet-domain representations $Q_R^W$ and $Q_I^W$ so that their inverse wavelet transforms $Q_R$ and $Q_I$ satisfy (3). To begin, the input image $S \in \mathbb{H}^{H \times W}$ is used to form two initial quaternions $Q_R^{init} \in \mathbb{H}^{H \times W}$ and $Q_I^{init} \in \mathbb{H}^{H \times W}$ using the normalized and original quaternion color representations (4) and (6). Each quaternion $Q_R^{init}$ or $Q_I^{init}$ consists of four channels $(T_w, T_x, T_y, T_z)$, where $T_w$ is the real part and $T_x, T_y$, and $T_z$ are the three imaginary parts. A discrete Haar wavelet transform $\mathcal{W}\{\cdot\}$ is applied independently to every channel, yielding wavelet-domain tensors:

$$Q_R^W = (\mathcal{W}\{T_w\}, \mathcal{W}\{T_x\}, \mathcal{W}\{T_y\}, \mathcal{W}\{T_z\}) \text{ and}$$

$$Q_I^W = (\mathcal{W}\{T'_w\}, \mathcal{W}\{T'_x\}, \mathcal{W}\{T'_y\}, \mathcal{W}\{T'_z\})$$

where $(T'_w, T'_x, T'_y, T'_z)$ denote the channels of $Q_I^{init}$. Concatenating these results produces a 32-channel representation, where the 16 channels of $Q_R^W$ and 16 channels of $Q_I^W$ capture multiscale information in the LL, LH, HL, and HH sub-bands.

Once in the wavelet domain, the network refines the representations through separate convolutional branches that learn higher-level features for each quaternion. A Symmetric Cross Attention module is employed to enforce coherent decomposition.

**Cross Attention module:** Let the wavelet-domain feature map for the reflectance branch be $q_r \in \mathbb{R}^{C \times H \times W}$ and for the illumination branch be $q_i \in \mathbb{R}^{C \times H \times W}$. A learnable set of pointwise convolutions projects $q_r$ and $q_i$ into query $Q_r$, key $K_i$, and value $V_i$ tensors. Reflectance branch computes (the inner working of the block is illustrated in Fig 3.):

$$Q_r = W_{qr} * q_r, \quad K_i = W_{ki} * q_i, \quad V_i = W_{vi} * q_i$$

Illumination branch computes:

$$Q_i = W_{qi} * q_i, \quad K_r = W_{kr} * q_r, \quad V_r = W_{vr} * q_r$$

Each query, key, or value is reshaped so that the spatial dimensions (H × W) become a single index, and let d be the per-head dimensionality. Then cross-attention for $q_r$ is defined as:

$$A_r = \text{softmax}\left(\frac{Q_r}{K_i^T \sqrt{d}}\right), \quad \text{cross}_r = A_r V_i$$

here $A_r$ is the attention matrix, and $\text{cross}_r$ is the resulting aggregation from the illumination branch back into $q_r$. A similar cross-attention is formed to refine $q_i$:

$$A_i = \text{softmax}\left(\frac{Q_i}{K_r^T \sqrt{d}}\right), \quad \text{cross}_i = A_i V_r$$

These cross-branch features are reshaped back to match the original spatial dimensions and fused with a residual skip-connection:

$$\hat{q}_r = q_r + W_o(\text{cross}_r), \quad \hat{q}_i = q_i + W_o(\text{cross}_i)$$

Where $W_o$ is a learned 1 × 1 convolution consolidating the multi-head outputs. This enables each branch to attend to features in the other, strengthening the decomposition of shared edges and color dependencies.

After cross-attention, the refined features $\hat{q}_r$ and $\hat{q}_i$ are concatenated and passed through a 1 × 1 convolution, reducing channel dimensionality, and fusing the information into a single tensor. To counteract blurring effects, a spatial sharpening step is applied. It is implemented by convolving with a Laplacian-like kernel that highlights edges and high-frequency details; its output is added back to the main signal to restore local contrast. An inverse Haar transform is then applied to each output channel, recovering the original spatial resolution while retaining the advantage of multiscale wavelet features. The resulting eight-channel tensor is split into two four-channel quaternions, $Q_r$ and $Q_i$ Finally, a smoothing block is applied to each quaternion to reduce grid artifacts that arise from wavelet decomposition and reconstruction. The overall strategy preserves edges, maintains color fidelity, and enables a robust factorization of reflectance and illumination by Retinex principles.

*C. Implementation details*

The proposed quaternion-based decomposition network is developed in PyTorch and optimized with AdamW optimizer [60], [61]. To stabilize training and to prevent overfitting, we employ a two-stage learning rate schedule: (i) a linear warm-up from an initial rate of $3 \times 10^{-4}$ over the first 10 epochs, followed by (ii) a cosine annealing phase that gradually decreases the learning rate to $1 \times 10^{-7}$ by epoch 1000. We train on 256 × 256 image patches.

We define a multi-term loss function $\mathcal{L}_{decom}$ to simultaneously enforces accurate reconstruction from both low-light and high-light images, maintains consistency under different illuminations, preserves smoothness of illumination to avoid halo artifacts. The total decomposition loss is defined as:

$$\mathcal{L}_{decom} = \underbrace{\mathcal{L}_{\text{recon\_low}}}_{(a)} + \underbrace{\mathcal{L}_{\text{recon\_high}}}_{(b)} + 0.01 \underbrace{\mathcal{L}_{\text{recon\_mutual\_low}}}_{(c)}$$
$$+ 0.01 \underbrace{\mathcal{L}_{\text{recon\_mutual\_high}}}_{(d)} + 0.05 \underbrace{\mathcal{L}_{\text{smooth}}}_{(e)} + 0.01 \underbrace{\mathcal{L}_{\text{equal\_R}}}_{(f)}$$
$$+ 0.01 \underbrace{\mathcal{L}_{\text{freq}}}_{(g)}$$

**(a) & (b) Reconstruction losses:** We denote the quaternion-valued reflectance and illumination for low-light input by $Q_R^{low}$, $Q_I^{low}$, and normal light input by $Q_R^{high}$, $Q_I^{high}$. The reconstructed images $\hat{S}_{low}$ and $\hat{S}_{high}$ are obtained from the imaginary parts of their Hamilton products:

$$\hat{S}_{low} = Q_R^{low} \otimes Q_I^{low}, \quad \hat{S}_{high} = Q_R^{high} \otimes Q_I^{high}$$

The reconstruction losses are defined as:
$$\mathcal{L}_{\text{recon\_low}} = \|\hat{S}_{low} - S_{low}\|_1$$
$$\mathcal{L}_{\text{recon\_high}} = \|\hat{S}_{high} - S_{high}\|_1$$

where $\|\cdot\|_1$ is the $\ell_1$ norm, promoting less blur compared to $\ell_2$. $S_{low}$ and $S_{high}$ denote input low-light and normal-light images, respectively. These terms ensure that the reflectance–illumination decomposition accurately reconstructs both low-light images and ground-truth high-light images.

**(c) & (d) Mutual reconstruction losses:** To stabilize decomposition under incomplete or mismatched reflectance–illumination pairs, we also consider cross-reconstructions:

$$\hat{S}_{low} = Q_R^{low} \otimes Q_I^{low} \quad , \quad \hat{S}_{high} = Q_R^{high} \otimes Q_I^{high}$$

We impose:
$$\mathcal{L}_{\text{recon\_mutual\_low}} = \|\hat{S}_{\text{mutual\_low}} - S_{low}\|_1$$
$$\mathcal{L}_{\text{recon\_mutual\_high}} = \|\hat{S}_{\text{mutual\_high}} - S_{high}\|_1$$

These terms limit degenerate solutions and encourage a more robust latent factorization that remains consistent under varying illuminations.

**(e) Illumination Smoothness:** To ensure that illumination varies smoothly across an image while preserving reflectance details, we define the illumination smoothness function as:

$$SM = \sum_{u \in \{x,y\}} |\nabla_u I_{\text{gray}}|_1 \odot \exp(-10 \nabla_u R_{\text{gray}}),$$

where $\nabla_u$ denotes finite differences in direction $u$, $I_{\text{gray}}$ and $R_{\text{gray}}$ are grayscale versions of illumination and reflectance, respectively, and $\odot$ is elementwise multiplication. This enforces smooth illumination variations with exception of the regions where strong reflectance edges are present, reducing the halo artifacts.

The total illumination smoothness loss is defined as:
$$\mathcal{L}_{\text{smooth}} = SM(Q_I^{low}, Q_R^{low}) + SM(Q_I^{high}, Q_R^{high}).$$

**(f) Reflectance Consistency:** To enforce consistency between reflectances estimated from low- and high-light images, we introduce the reflectance consistency loss:

$$\mathcal{L}_{\text{equal\_R}} = \|Q_R^{low} - Q_R^{high}\|_1$$

This term ensures reflectance stability under varying illuminations, enhancing decomposition robustness.

**(g) Frequency regularization:** To prevent the network from introducing undesired high-frequency noise in the final reconstructions, we apply a Fourier-domain regularization loss:

$$\mathcal{L}_{\text{freq}} = \gamma \mathbb{E}_{x \sim \mathcal{D}} \left[ |\mathcal{F}\{\hat{S}_{low}\}|_{HF} + |\mathcal{F}\{\hat{S}_{height}\}|_{HF} \right]$$

Here, $\mathcal{F}$ denotes the 2D Fourier Transform, $|\cdot|_{HF}$ isolates high-frequency components, $\gamma$ is a weighting factor (0.01 in our experiments), $\mathbb{E}_{x\sim\mathfrak{D}}[\cdot]$ denotes expectation over input samples $x$ drawn from dataset $\mathfrak{D}$. This loss prevents over-sharpening and ensures that reconstructed images remain visually coherent without artificial high-frequency artifacts.

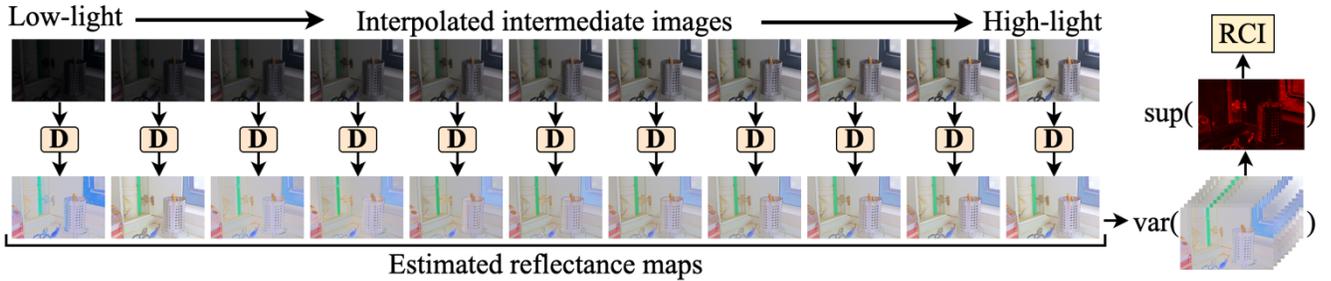

**Fig 4.** RCI Metric Computation. We linearly interpolate between the low-light and normal-light images to generate intermediate images $\{S_\alpha\}$. Each $S_\alpha$ is decomposed by $D$ to obtain a reflectance map $R_\alpha$. The pixel-wise variance of $\{R_\alpha\}$ is computed across $\alpha$, and its supremum over all pixels is normalized by the maximum possible variance. The final RCI indicates how robust the reflectance is to illumination changes

*D. The RCI metric*

To assess the stability of a decomposition $D$ under varying illumination, we introduce the Reflectance Consistency Index (RCI). We begin by linearly interpolating between a low-light image $S_{low}$ and a normal-light image $S_{low}$, yielding a sequence $\{S_\alpha\}$ for $\alpha \in [0,1]$:

$$S_\alpha = (1-\alpha)S_{low} + \alpha S_{normal}$$

Applying the chosen decomposition D to each $S_\alpha$ produces reflectance maps $\{R_\alpha(x,y)\}$. If the reflectance captures intrinsic scene properties, it should remain nearly invariant across different illumination conditions.

Let $\Omega$ denote the set of all pixel locations $(x,y)$. For each pixel, we compute the variance of its reflectance values $\{R_\alpha(x,y)\}$ over $\alpha \in [0,1]$:

$$\sigma_\alpha^2\big(R_\alpha(x,y)\big) = \frac{1}{|\mathcal{A}|}\sum_{\alpha\in\mathcal{A}}\big(R_\alpha(x,y)\big)^2 - \left(\frac{1}{|\mathcal{A}|}\sum_{\alpha\in\mathcal{A}}R_\alpha(x,y)\right)^2$$

where $\mathcal{A} \subseteq [0,1]$ is the discrete set of interpolation steps. Since the reflectance is assumed normalized to the range $[0,1]$, the maximum possible variance is $\sigma_{max}^2 = 0.25$. We then define the RCI by taking the supremum of these variances over $\Omega$:

$$\text{RCI} = 1 - \frac{\sup_{x\in\Omega}\sigma_\alpha\big(R_\alpha(x)\big)}{\sigma_{max}^2}$$

An RCI value of 1 corresponds to zero variance and indicates perfectly consistent reflectance across all illumination levels. Lower values signify greater deviations in $R_\alpha$. This worst-case perspective ensures that even localized inconsistencies in the reflectance map are penalized. By focusing on the supremum of the variance, RCI captures the most critical breakdown in reflectance invariance. This ensures that methods yielding highly stable reflectance under most pixels, yet large errors in a few regions, are penalized, reflecting the core Retinex principle that the reflectance should remain consistent under changes in illumination.

## IV. EXPERIMENTS

We evaluate the proposed quaternion-based Retinex decomposition across diverse low-light vision tasks, including low-light crack detection, zero-shot day-night adaptation for object detection, and infrared-visible image fusion. We compare our approach with existing state-of-the-art methods, analyzing both decomposition accuracy and downstream task performance.

*A. Decomposition Experiments*

We first assess our quaternion-based Retinex decomposition on the LOLv1 dataset [33], which contains 500 pairs of low-light and normal-light images. Our evaluation focuses on two core aspects: (i) the numerical accuracy of the reflectance and illumination reconstruction, and (ii) the visual quality and consistency of the reflectance component under varying illumination. We use PSNR (dB) and SSIM to measure fidelity and structural similarity, LPIPS to measure perceptual quality, and the newly proposed RCI to quantify how stable the reflectance remains across different illuminations. Table I summarizes the reconstruction performance of our method and several state-of-the-art approaches. The competing methods do not explicitly optimize for exact back-reconstruction, leading to noticeable discrepancies and artifacts in the reflectance. By contrast, our approach fully leverages cross-channel relationships, yielding consistently higher PSNR and SSIM scores and a markedly improved RCI. The substantial RCI gain indicates that our reflectance remains far more invariant when the illumination changes, highlighting the strength of modeling color channels jointly via quaternion operations.

TABLE I
RECONSTRUCTION ACCURACY

| Decomposition | Architecture | PSNR | SSIM | LPIPS | RCI |
|---|---|---|---|---|---|
| RetinexNet [33] | CNN | 44.807 | 0.9723 | 0.6119 | 0.779 |
| KIND++[34] | CNN | 45.941 | 0.9795 | 0.0373 | 0.605 |
| URetinexNet [35] | CNN | 36.131 | 0.8549 | 0.5381 | 0.824 |
| Diff-Retinex [36] | CNN+Transformer | 46.309 | 0.9795 | 0.0354 | 0.774 |
| Our | CNN | **62.515** | **0.9998** | **0.0001** | **0.988** |

Figures 5–6 offer a closer look at typical decomposition outputs. RetinexNet [33] (Fig. 5b) tends to amplify noise and produce edge halos, while KIND++ [34] (Fig. 5d) reduces noise but often loses fine texture details in the reflectance. URetinex and Diff-Retinex [35], [36] (Fig. 5d and 5e) better suppress noise but introduce mild color shifts around high-contrast edges. By contrast, our method (Fig. 5f) effectively preserves both texture and color fidelity.

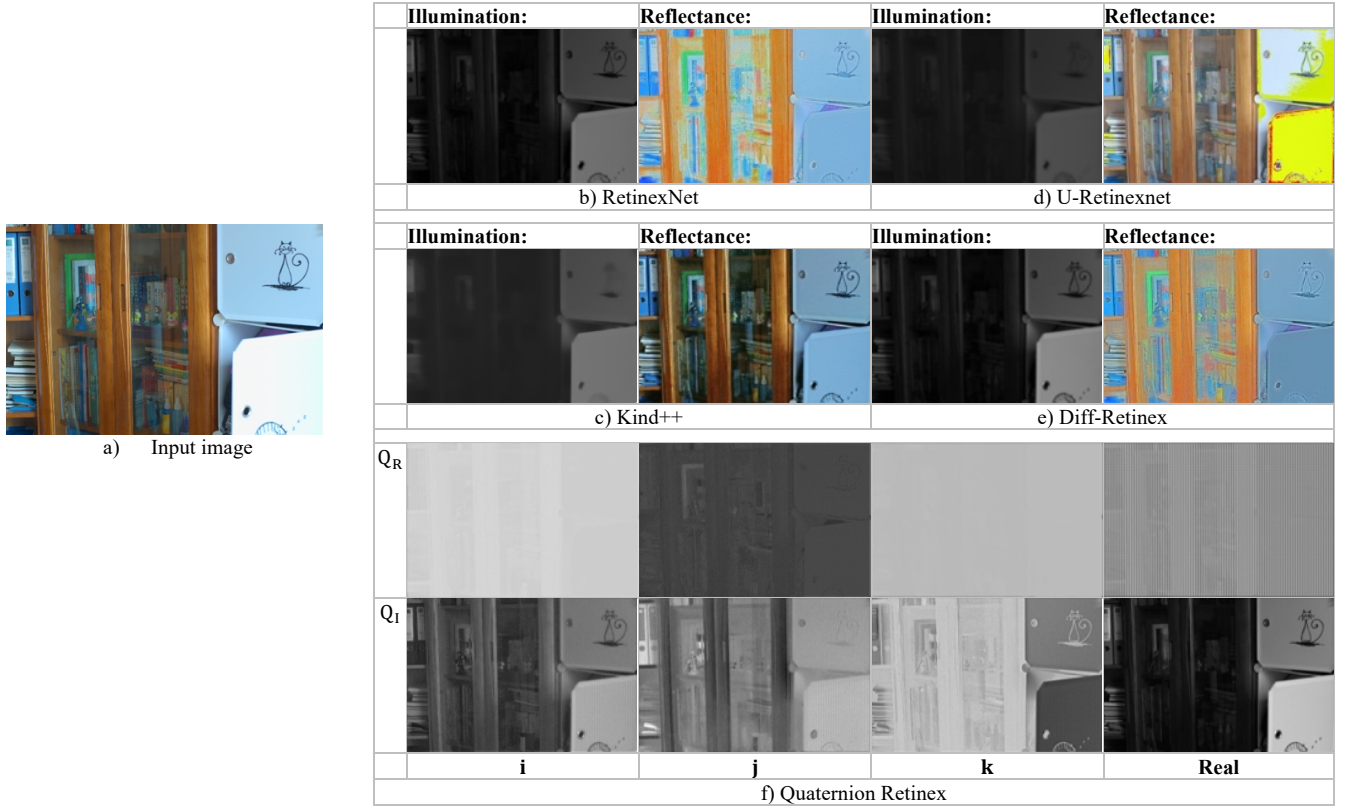

**Fig. 5.** Retinex decomposition (illumination-reflectance) by deep-learning based approaches. Notice the halos and color shifts in the reflectance maps. *a)* Input image *b)* RetinexNet c) Kind++ d) U-Retinexnet e) Diff-retinex, f) Quaternion Retinex. Our decomposition results for the same images, showing more consistent reflectance and smoother illumination

**Fig. 6.** Variance of the reflectance map with change of the illumination for various methods.

The reflectance component retains structural details, and the illumination remains smooth without introducing artifacts. Notably, our reflectance remains stable even under simulated illumination changes, which is well reflected by its high RCI score. These results confirm that the proposed quaternion-based Retinex framework is more adept at capturing cross-channel dependencies while ensuring perfect reconstruction. The higher PSNR and SSIM metrics, in tandem with the lower LPIPS and higher RCI, underscore the robustness and consistency of our decomposition approach, an important advantage for downstream tasks in low-light imaging.

### B. Low-light crack segmentation

Detecting and segmenting cracks in concrete structures is critical for structural health monitoring, yet such tasks are often complicated by poor visibility, noise, and uneven illumination. CrackNex approaches the problem by using a Retinex-based decomposition to extract illumination-invariant reflectance features and few-shot learning to deal with the scarcity of labeled crack images [62]. In the original CrackNex pipeline, an input image $S \in \mathbb{R}^{H \times W \times 3}$ is decomposed into reflectance and illumination components $(R, I)$.

TABLE II

| Method | mIoU (%) | F1-score (%) |
|---|---|---|
| CrackNex (original) | 65.6 | 72.3 |
| Quaternion (ours) | 67.6 | 75.4 |
| Wavelet (ours) | **68.7** | **76.2** |

A few-shot support prototype is then learned from limited labeled exemplars, while a reflectance prototype is extracted from R. These prototypes are integrated into a dedicated module to simultaneously include crack features and reflectance information, followed by an Atrous Spatial Pyramid Pooling (ASPP) module to enhance multi-scale features for final segmentation. The original implementation uses a pretrained network from RetinexNet for decomposition [33]. We replace it with our decomposition, where the RGB channels S are considered a single quaternion entity. This quaternion $S_q$ is then decomposed into quaternion-valued reflectance $Q_R$ and illumination $Q_I$. Formally $S_q = Q_R \otimes Q_I$.

We evaluate two variations: (i) Standard quaternion-based decomposition in the spatial domain, and (ii) the output of the decomposition network before the inverse wavelet transform. Only the reflectance component $Q_R$ is used to learn the reflectance prototype. All other components of *CrackNex*, including few-shot prototype learning, prototype fusion, and ASPP segmentation, remain intact.

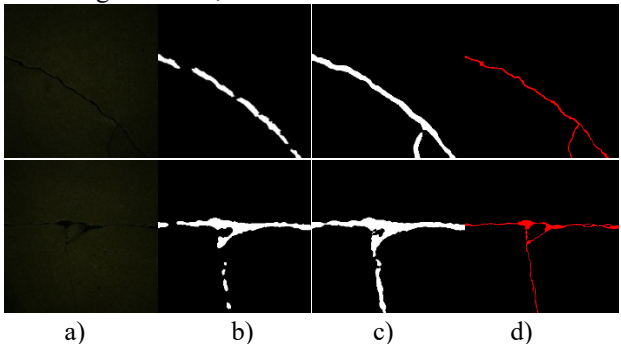

a)     b)     c)     d)

**Fig. 7.** Low-light crack segmentation. a) Input b) *CrackNex* c) Our method d) Ground truth

The primary difference is that the scalar reflectance constraint replaces our quaternion-based constraint.

We benchmark the modified *CrackNex* on the LCSD dataset. As summarized in Table II, the quaternion-based methods achieve superior segmentation results: our spatial-domain Quaternion Retinex yields a +2% boost in mean Intersection-over-Union (mIoU) and a +3% increase in F1-score compared to the original *CrackNex*. When further incorporating the wavelet-domain decomposition, we observe an even larger improvement. Qualitative results in Fig. 7 illustrate that our Quaternion Retinex significantly enhances crack boundary visibility and suppresses noise in dark regions.

*C. Zero-Shot Day-Night Domain Adaptation for Object Detection*

Object detectors trained on well-lit, daytime images often struggle under low-light or night-time conditions due to the considerable domain gap created by poor illumination, high noise, and color distortions. Assembling large-scale labeled datasets for night scenes is costly, which drives the development of zero-shot domain adaptation approaches to enable generalization to nighttime without needing actual nighttime data during training. Du et al. proposed a "DArk-Illuminated Network" (DAI-Net) that pairs normal-light images with their artificially darkened counterparts [63]. This work also utilizes the decomposition from the Retinex module, thereby directing the model to concentrate on reflectance features that are less influenced by changes in illumination. To enhance these features, DAI-Net employs a loss function that encourages consistent reflectance estimates across synthetic and real variations in illumination. We replace RetinexNet with our Quaternion Retinex. Concretely, the synthetic low-light image $S_q^{(syn)}$ and its well-lit counterpart $S_q^{(lit)}$ are jointly decomposed into quaternion reflectance $Q_R$ and illumination $Q_I$. We adapt DAI-Net's interchange and recomposition steps to operate on $Q_R$ and $Q_I$, applying the same reconstruction losses in quaternion space. Apart from the decomposition module, the downstream detection pipeline (e.g., YOLOv3 or DSFD) is left unchanged.

We evaluate our approach on Dark Face dataset under zero-shot adaptation settings [64]. Table III compares the original *DAI-Net* with our *DAI-Net* + quaternion retinex variant, reporting mean Average Precision (mAP) on two benchmark datasets. Integrating quaternion retinex boosts performance by +1.2% mAP on the Wider-Face → Dark-Face setup and +0.8% mAP on the COCO → Dark Face transfer.

TABLE III

| Method | Wider-Face → DarkFace (mAP %) | CoCo → DarkFace (mAP %) |
|---|---|---|
| DAI-Net | 52.9 | 57.0 |
| DAI-Net + Quaternion Retinex | **54.1** | **57.8** |

*D. Retinex-Based Infrared and Visible Image Fusion*

Infrared-visible (IR-VI) fusion is crucial for night vision, surveillance, and target recognition. IR images capture thermal information while visible images provide texture and color details. Recent methods use Retinex decomposition to separate visible image reflectance from illumination, then fuse with infrared data at feature or pixel levels. Wang et al. [66] propose RDMFuse, an original decomposition model independent of pretrained networks. It fuses infrared with visible reflectance to preserve structural details, using an illumination-adaptive module to enhance contrast in low-light conditions. We have replaced the standard reflectance-illumination separation with our quaternion-based wavelet-domain decomposition. Particularly:

1. We decompose the visible image using our quaternion wavelet model, yielding $Q_R^{VI}$ and $Q_I^{VI}$ in the quaternion domain.
2. We convert the infrared image into a quaternion representation $Q^{IR}$ by assigning its intensity values to the imaginary components of the quaternion and setting the real component to zero.
3. The existing contrast texture module and reflectance-fusion function in *RDMFuse* now operate on $Q_R^{VI}$ and $Q^{IR}$.
4. The illumination-adaptive module remains in place but processes our quaternion-based illumination $Q_I^{VI}$ instead of the original scalar illumination map. Minor normalization layers are introduced to handle the four-dimensional hypercomplex features.

We evaluated our approach on the LLVIP dataset using four visual quality metrics: EN (Entropy): Information content; higher values indicate richer detail SF (Spatial Frequency): Overall sharpness; larger values imply sharper images AG (Average Gradient): Edge clarity and contrast; higher values indicate better edge detail STD (Standard Deviation): Intensity spread; higher values indicate better contrast.

TABLE IV

| Method | EN | SF | AG | STD |
|---|---|---|---|---|
| RDMFuse | 6.52 | 12.48 | 3.21 | 24.8 |
| Quaternion Retinex | 6.70 | 12.89 | 3.34 | 25.1 |

Table IV reports the average metric scores on the LLVIP test subset. Our method exceeds *RDMFuse* across all four metrics, suggesting improved detail preservation, edge clarity, and contrast. Hence, quaternion retinex demonstrates more robust low-light fusion performance than RDMFuse, aided by quaternion-based wavelet decomposition and hypercomplex illumination handling.

*E. Ablation studies*

TABLE V
ABLATION STUDY

| | WT | CA | FR | PSNR | SSIM | MSE |
|---|---|---|---|---|---|---|
| Baseline | ✗ | ✗ | ✗ | 51.12 | 0.881 | 0.0213 |
| +WT | ✓ | ✗ | ✗ | 53.08 | 0.891 | 0.0192 |
| +WT+CA | ✓ | ✓ | ✗ | 54.74 | 0.908 | 0.0183 |
| +WT+CA+FR | ✓ | ✓ | ✓ | 62.51 | 0.999 | 0.0171 |

An ablation study was conducted to evaluate the impact of key design components on reflectance reconstruction: the wavelet-domain transform (WT), the symmetric cross-attention mechanism (CA), and frequency regularization (FR). The experiments were performed on the LOLv1 dataset, and the results are shown in Table V. The table compares four model variants in terms of average PSNR and SSIM for the reconstructed reflectance and the mean squared error on the reflectance maps. Introduction of the wavelet-domain transform reduces high-frequency noise and increases PSNR and SSIM. Enabling symmetric cross-attention further refines reflectance by mitigating the leakage of illumination cues, leading to fewer edge artifacts. Incorporating frequency regularization yields an additional 0.5 dB improvement in PSNR and slightly lower MSE, suggesting that restricting high-frequency noise is key to stable, high-quality decomposition.

## V. CONCLUSIONS AND FUTURE WORK

This paper presents a wavelet-domain quaternion-based Retinex decomposition that addresses traditional scalar methods' limitations by representing RGB channels as unified hypercomplex entities. Our approach captures cross-channel dependencies through quaternion algebra and Haar wavelets, ensuring perfect reconstruction via $S=R\otimes I$. The method achieves superior performance with PSNR up to 62.51 dB and SSIM of 0.9998 on LOLv1. The novel Reflectance Consistency Index (RCI) reaches 0.988, validating reflectance stability under challenging illumination. Across applications including low-light crack detection, nighttime face detection, and infrared-visible fusion, our framework consistently outperforms state-of-the-art methods by 2-11%. The quaternion framework preserves fine-grained color information while suppressing noise, maintaining cross-channel correlations that scalar methods ignore. This ensures enhanced color fidelity and superior structural preservation in challenging low-light scenarios.

Key directions for future work include developing perceptual enhancement frameworks that incorporate opponent color processing and CIE-aligned quaternion representations, extending to temporal modeling for video sequences with quaternion-based consistency constraints, and optimizing computation through GPU-accelerated kernels and lightweight architectures for mobile deployment. Additionally, aligning algorithmic parameters with human visual system properties will improve interpretability across diverse conditions, while addressing robustness to extreme lighting and ensuring generalization across different imaging systems remain critical research opportunities. The proposed framework provides a strong foundation for advancing low-light image enhancement with both theoretical and practical performance improvements.